\documentclass{article}
\usepackage[T1]{fontenc}
\usepackage{iclr2027_conference,times}
\usepackage{amsmath,amssymb,amsthm,mathtools}
\usepackage{graphicx,booktabs,tabularx,microtype}
\usepackage{colortbl}   %
\usepackage{hyperref,url,xcolor}
\usepackage{enumitem}
\usepackage{placeins}
\usepackage{listings}
\definecolor{codebackground}{HTML}{F5F7F9}
\definecolor{codeaccent}{HTML}{285577}
\definecolor{codemuted}{HTML}{73808C}
\lstdefinestyle{paperpython}{
  language=Python,
  deletekeywords={len,sum},
  basicstyle=\fontsize{9.5}{12.5}\selectfont\ttfamily,
  keywordstyle=\color{codeaccent}\bfseries,
  stringstyle=\color{codeaccent},
  emph={train_temporal,train_dynamics,choose_action},
  emphstyle=\bfseries,
  backgroundcolor=\color{codebackground},
  frame=l,
  rulecolor=\color{codeaccent},
  framerule=1pt,
  framesep=8pt,
  xleftmargin=22pt,
  xrightmargin=8pt,
  numbers=left,
  numberstyle=\scriptsize\color{codemuted},
  numbersep=10pt,
  columns=fullflexible,
  keepspaces=true,
  showstringspaces=false,
  breaklines=true,
  aboveskip=10pt,
  belowskip=10pt
}
\usepackage[capitalise]{cleveref}
\crefname{figure}{Fig.}{Figs.}                \Crefname{figure}{Figure}{Figures}
\crefname{table}{Tab.}{Tabs.}                 \Crefname{table}{Table}{Tables}
\crefname{equation}{Eq.}{Eqs.}                \Crefname{equation}{Equation}{Equations}
\crefname{section}{Sec.}{Secs.}               \Crefname{section}{Section}{Sections}
\crefname{subsection}{Sec.}{Secs.}            \Crefname{subsection}{Section}{Sections}
\crefname{subsubsection}{Sec.}{Secs.}         \Crefname{subsubsection}{Section}{Sections}
\crefname{appendix}{App.}{Apps.}              \Crefname{appendix}{Appendix}{Appendices}
\crefname{subappendix}{App.}{Apps.}           \Crefname{subappendix}{Appendix}{Appendices}
\crefname{subsubappendix}{App.}{Apps.}        \Crefname{subsubappendix}{Appendix}{Appendices}
\crefname{proposition}{Prop.}{Props.}         \Crefname{proposition}{Proposition}{Propositions}
\crefname{corollary}{Cor.}{Cors.}             \Crefname{corollary}{Corollary}{Corollaries}
\crefname{lemma}{Lem.}{Lems.}                 \Crefname{lemma}{Lemma}{Lemmas}
\crefname{paperalgorithm}{Alg.}{Algs.}        \Crefname{paperalgorithm}{Algorithm}{Algorithms}
\newcolumntype{C}{>{\centering\arraybackslash}X}   %
\definecolor{customlinkcolor}{HTML}{2774AE}
\definecolor{customcitecolor}{HTML}{2774AE}
\hypersetup{
  colorlinks=true,
  linkcolor=customlinkcolor,
  anchorcolor=black,
  citecolor=customcitecolor,
  urlcolor=customlinkcolor
}
\newcommand{\E}{\mathbb E}
\DeclareMathOperator*{\argmax}{arg\,max}
\newtheorem{proposition}{Proposition}

\iclrfinalcopy

\title{Learning to Plan from Random Exploration}
\makeatletter
\newcommand{\paperauthorblock}{%
\parbox[t]{\dimexpr\textwidth-2\tabcolsep\relax}{%
\raggedright
{\normalfont
\renewcommand{\@makefnmark}{\hbox{\@textsuperscript{\normalfont\@thefnmark}}}
\textbf{Deqian Kong}\textsuperscript{1,}\thanks{Equal contribution.},\hspace{0.1em}
\textbf{Guangyan Sun}\textsuperscript{2,}\footnotemark[1],\hspace{0.1em}
\textbf{Sheng Cheng}\textsuperscript{3,}\footnotemark[1],\hspace{0.1em}
\textbf{Sirui Xie}\textsuperscript{1},\hspace{0.1em}
\textbf{Bo Pang}\textsuperscript{4},\\[0.3ex]
\textbf{Jianwen Xie}\textsuperscript{5},\hspace{0.1em}
\textbf{Tony Geng}\textsuperscript{6},\hspace{0.1em}
\textbf{Caiwen Ding}\textsuperscript{2},\hspace{0.1em}
\textbf{Ying Nian Wu}\textsuperscript{1}\par}
\vspace{0.7ex}
{\normalfont\small
\mbox{\textsuperscript{1}UCLA\enspace
\textsuperscript{2}University of Minnesota\enspace
\textsuperscript{3}Amazon AGI\enspace
\textsuperscript{4}Salesforce Research\enspace
\textsuperscript{5}Lambda\enspace
\textsuperscript{6}Rice University}\par}
}}
\makeatother
\author{\paperauthorblock}

\begin{document}
\makeatletter
\begingroup
\renewcommand{\@fnsymbol}[1]{\ensuremath{\star}}
\maketitle
\endgroup
\makeatother
\lhead{Preprint. Work in Progress.}

\begin{abstract}
Random exploration reveals how an environment can be traversed
before a goal is specified. Can this experience support long-range
planning without policy-improvement training?
Our random-walk analysis explains what temporal relations contain:
short horizons reveal geodesic geometry in the diffusion limit,
while longer horizons reveal connectivity between regions before
mixing removes these distinctions.
We learn these relations with a conditional energy-based model
that estimates temporal log-density ratios through horizon-conditioned
embeddings. The model is trained on observation pairs by
noise-contrastive estimation, without action or reward labels.
The planner queries these learned relations at different horizons
as it moves toward the goal.
At test time, a separate local dynamics model predicts candidate
action outcomes, and the temporal model evaluates their progress
toward the goal by selecting or aggregating estimated improvements
across horizons. The agent executes one action and replans with
both models fixed.
Experiments demonstrate long-range maze planning from random
exploration using states and images. Learned score fields,
embedding probes, and planned routes exhibit properties of a
multiscale cognitive map. We further demonstrate egocentric
navigation from random exploration and manipulation planning
from suboptimal data.
\end{abstract}

\addtocontents{toc}{\protect\setcounter{tocdepth}{-1}}
\section{Introduction}
\label{sec:intro}

In a classic latent-learning experiment, rats explored a complex maze for ten
days with no food waiting at the goal
\citep{tolman1930introduction}. After food was introduced, rats that
had wandered unrewarded took nearly as few wrong turns on subsequent
trials as rats rewarded from the start. Exploration had taught them
far more than their behavior had revealed. The cognitive-map
hypothesis and studies of hippocampal place cells link navigation to
internal spatial representations
\citep{tolman1948cognitive,okeefe1971spatial,zhao2025placecells}, while sequence-learning
accounts explain how these representations can arise from temporal
experience \citep{raju2024space}. We ask whether even
\emph{random} exploration is enough: can representations learned from
trajectories collected without goal-directed or curiosity-driven
action selection support long-range planning?

A random walk through an environment leaves temporal traces at
multiple scales. Observations a few steps apart reveal local
connections between states, like reading a map at street level.
Observations many steps apart reveal connections between regions,
like reading it at district level. We show that, under random-walk
assumptions, short-horizon statistics recover geodesic geometry in
the diffusion limit while longer horizons capture regional
connectivity before mixing erases these distinctions. Classical
results on heat kernels and diffusion processes
\citep{varadhan1967heat,norris1997heat,coifman2006diffusion}
provide the basis for this analysis. The challenge is to learn these
relations from sampled trajectories, without access to the full
transition kernel, and let a planner query different scales as it
moves toward a goal.

Existing methods obtain long-range guidance in different ways.
Goal-conditioned reinforcement learning learns values and trains
policies to act on them \citep{eysenbach2022crl,park2023hiql}.
Map-like representations can emerge through navigation objectives
\citep{wijmans2023emergence}. Latent world models learn
action-conditioned predictions from reward-free trajectories and
evaluate multistep action sequences at test time
\citep{zhou2025dinowm,wang2026temporal}, but their planners depend on
rollout accuracy and a latent geometry that reflects goal progress.
Temporal contrastive representations can encode probability ratios
that support planning via interpolation in representation space
\citep{eysenbach2024interpolation}. We separate long-range guidance
from local dynamics prediction: a representation learned across
horizons evaluates progress toward the goal, while a one-step
dynamics model predicts the outcomes of candidate actions.
Planning requires neither a trained goal-reaching policy nor
multistep dynamics rollouts.

We model temporal relations with a conditional energy-based model~\citep{lecun2006tutorial,xie2016generative}
whose source and target embeddings estimate temporal log-density
ratios at each horizon. Noise-contrastive estimation~\citep{gutmann2010nce} learns them from
observation pairs, requiring no action or reward labels. At test
time, the dynamics model predicts immediate action outcomes and the
temporal representation scores their estimated improvement in goal
likelihood. The temporal representation is reusable across goals,
and the agent replans after each action with both models fixed.

Our contributions are threefold:
\begin{enumerate}[label=(\arabic*), leftmargin=*, nosep]
    \item We develop a conditional temporal model that learns
    multi-horizon representations from exploration trajectories and
    supports closed-loop goal reaching without policy-improvement
    training.

    \item We characterize the geodesic and connectivity information
    in the ideal random-walk temporal score, and unify greedy horizon
    selection and potential ascent through a temperature-controlled
    planning objective.

    \item We demonstrate long-range maze planning from random
    exploration using states and images. Score fields, embedding
    probes, and planned routes reveal properties of a multiscale
    cognitive map. Further evaluations cover egocentric navigation
    and manipulation with suboptimal data.
\end{enumerate}

\section{Method}
\label{sec:model}
Exploration trajectories reveal how an environment can be traversed before a goal is specified. We learn these temporal relations at multiple horizons, then query them to turn predictions of local action outcomes into goal-directed decisions.

\subsection{Problem Formulation}
Given a dataset of trajectories in which each observation $x_t$ is a state vector or an image, we aim to learn temporal relations that support goal-directed planning. Our focus is random exploration for navigation: locally feasible moves are sampled without reference to a task goal. The learning formulation also accepts suboptimal or expert trajectories, which can be useful in manipulation tasks. Data collection can therefore balance broad exploration with coverage of task-relevant transitions. In each case, we learn the temporal relations induced by the behavior represented in the dataset.

Let $\mathcal T$ be a finite set of positive prediction horizons. At horizon $\tau\in\mathcal T$, we sample a source--target pair $(x,y)=(x_t,x_{t+\tau})$ from the same episode, with distribution $p_{\rm data}(x,y|\tau)$. This distribution describes where exploration leads from $x$ after $\tau$ steps, and varying $\tau$ describes this at different time scales. We model $p_{\rm data}(y|x,\tau)$ so that, once a goal is given, the planner can ask how likely each candidate next observation is to lead to it within $\tau$ steps. Learning this conditional uses only observation pairs, without action or reward labels. Action-labeled one-step transitions $(x_t,a_t,x_{t+1})$ are used separately to learn local dynamics.
\subsection{Models}
\label{sec:model_setting}
\noindent\textbf{A multi-horizon representation.}
The temporal relation between two observations depends on how long the agent explores. We describe it through the alignment of a source embedding $h_\theta(x,\tau)$ and a target embedding $g_\theta(y,\tau)$. The family
$\{h_\theta(\cdot,\tau),g_\theta(\cdot,\tau)\}_{\tau\in\mathcal T}$
represents how these relations change with the horizon. Two observations may have little association over a short horizon but become associated once exploration has had time to traverse a connecting route.

The planner can query this family at different horizons to evaluate the same candidate move at several temporal scales. In spatial environments, these relations can exhibit properties of a \emph{multiscale cognitive map} \citep{tolman1948cognitive,zhao2025placecells}. \Cref{sec:theory} connects them to geodesic geometry and connectivity between regions under random-walk assumptions.

\noindent\textbf{Conditional temporal model.}
We model the target observation $y$ given the source $x$ and horizon $\tau$ with a conditional energy-based model (EBM). Let $p_0(y)$ be a reference distribution whose support covers the target observations. The model is
\begin{equation}
p_\theta(y|x,\tau)
=\frac{p_0(y)}{Z_\theta(x,\tau)}
\exp\!\left(
\beta_0(\tau)\langle h_\theta(x,\tau),g_\theta(y,\tau)\rangle
\right),
\label{eq:conditional_model}
\end{equation}
where
$
Z_\theta(x,\tau)=\E_{p_0(y)}\!\left[
\exp\!\left(\beta_0(\tau)\langle h_\theta(x,\tau),g_\theta(y,\tau)\rangle\right)
\right]
$
normalizes the conditional distribution. Both embeddings have unit norm,
$\|h_\theta(x,\tau)\|_2=\|g_\theta(y,\tau)\|_2=1$. For a given source and horizon, targets with larger alignment receive greater weight relative to $p_0$. The learned coefficient $\beta_0(\tau)$ controls the strength of this dependence across horizons.

\noindent\textbf{Parameterization.}
$h_\theta(x,\tau)$ and $g_\theta(y,\tau)$ are encoders that take an observation and a horizon as input. They receive a learned embedding of $\tau$, so one network covers all horizons and an observation can have a different embedding at each. The architecture depends on whether inputs are states or images. We write $\theta$ for all encoder, horizon-embedding, and coefficient parameters.

\noindent\textbf{Local dynamic model.}
The temporal model describes where exploration can lead. To evaluate an action, we also need a prediction of its immediate consequence. A separately parameterized local model predicts
\begin{equation}
\widehat x_{t+1}^{\,a}=F_\phi(x_t,a).
\label{eq:dynamics}
\end{equation}
The hat distinguishes this prediction from the observation obtained after execution. The temporal representation then evaluates the predicted outcome through its relations to the goal.

\subsection{Learning}
\label{sec:learning_algorithm}
\noindent\textbf{Noise-contrastive estimation.}
At a given horizon, we distinguish observed temporal pairs from pairs formed by sampling the target observation independently from $p_0$. These provide positive and negative examples, respectively. The EBM in \cref{eq:conditional_model} expresses temporal association through the log-density ratio
\begin{equation}
\begin{aligned}
\log\frac{p_\theta(y|x,\tau)}{p_0(y)}
&=\beta_0(\tau)\langle h_\theta(x,\tau),g_\theta(y,\tau)\rangle
-\log Z_\theta(x,\tau)\\
&\approx \beta_0(\tau)\langle h_\theta(x,\tau),g_\theta(y,\tau)\rangle+\beta_1(\tau)
=:G_\theta(x,y,\tau).
\end{aligned}
\label{eq:nce_model}
\end{equation}
When $p_0$ is the target marginal, the ratio compares how likely $y$ is to follow $x$ at horizon $\tau$ with how often $y$ occurs in the data. The learned offset $\beta_1(\tau)$ approximates $-\log Z_\theta(x,\tau)$, allowing us to fit $G_\theta$ by noise-contrastive estimation (NCE) without evaluating the partition function \citep{gutmann2010nce,ma2018conditionalnce}. Because the offset is shared across sources, source-wise normalization is approximate; \cref{app:calibration} details this distinction.

With $N$ negatives per positive and the same source marginal for both classes, the Bayes probability of a positive label is
\begin{equation}
\frac{p_{\rm data}(y|x,\tau)}
{p_{\rm data}(y|x,\tau)+N p_0(y)}
=\sigma\!\left(
\log\frac{p_{\rm data}(y|x,\tau)}{p_0(y)}-\log N
\right),
\label{eq:nce_discriminator}
\end{equation}
where $\sigma$ is the logistic sigmoid. Fitting $\sigma(G_\theta-\log N)$ therefore estimates the temporal log-density ratio. The reference enters through sampled negatives, so its density need not be evaluated.

We train across horizons by sampling $\tau$ from a chosen distribution $p(\tau)$ over $\mathcal T$ and minimizing
\begin{equation}
\begin{aligned}
\mathcal L_{\rm NCE}(\theta)
=\E_{p(\tau)}\!\Bigl[
&-\E_{p_{\rm data}(x,y|\tau)}\!\left[
\log\sigma\!\left(G_\theta(x,y,\tau)-\log N\right)
\right]\\
&-N\,\E_{p_{\rm data}(x|\tau)p_0(y)}\!\left[
\log\!\left(1-\sigma\!\left(G_\theta(x,y,\tau)-\log N\right)\right)
\right]
\Bigr].
\end{aligned}
\label{eq:nce}
\end{equation}
The two terms distinguish temporal pairs from independent pairs, using the same source marginal $p_{\rm data}(x|\tau)$. All temporal-model parameters are learned jointly across horizons. Our objective shares a similar loss form with SigLIP \citep{zhai2023siglip}.

At the unrestricted population optimum,
$G^*(x,y,\tau)=\log[p_{\rm data}(y|x,\tau)/p_0(y)]$
where both densities are positive. The constrained embedding model approximates this score, so $e^{G_\theta}$ estimates the conditional target density relative to the reference.

\noindent\textbf{Learning the local dynamics.}
We train the local predictor on action-labeled transitions at horizon $\tau=1$. Let $p_{\rm step}(x_t,a_t,x_{t+1})$ denote their empirical distribution. For state observations, we minimize
\begin{equation}
\mathcal L_{\rm dyn}(\phi)
=\E_{p_{\rm step}(x_t,a_t,x_{t+1})}\!\left[
\|F_\phi(x_t,a_t)-x_{t+1}\|_2^2
\right].
\label{eq:dynloss}
\end{equation}
The framework has a \emph{hierarchical} organization: local dynamics predicts one-step action outcomes, which the multi-horizon representation evaluates against the goal across horizons. Both are learned before planning and remain fixed at test time.

\subsection{Planning by Probability Improvement}
\label{sec:planning_algorithm}

Given a goal observation $y_g$, the planner favors actions whose predicted outcomes make the goal more likely. At the current observation $x_t$, the dynamics model predicts $\hat x^a_{t+1}=F_\phi(x_t,a)$ for each $a\in\mathcal A(x_t)$. We define progress at horizon $\tau$ as
\begin{equation}
\Delta_\tau(a)
:=e^{G_\theta(\hat x^a_{t+1},y_g,\tau)}
-e^{G_\theta(x_t,y_g,\tau)}.
\label{eq:horizon_progress}
\end{equation}
By \cref{eq:nce_model}, this quantity estimates the increase in goal probability relative to $p_0(y_g)$, $\Delta_\tau(a)\approx[p_\theta(y_g|\hat x^a_{t+1},\tau)-p_\theta(y_g|x_t,\tau)]/p_0(y_g)$.
At a fixed horizon, the current-state term is shared by all actions, so ranking actions by $\Delta_\tau(a)$ is equivalent to ranking their predicted outcomes by $e^{G_\theta(\hat x^a_{t+1},y_g,\tau)}$.

\noindent\textbf{Combining horizons.}
The same move can make different progress at different horizons. We combine these improvements with a temperature $\beta>0$, distinct from $\beta_0(\tau)$ and $\beta_1(\tau)$:
\begin{equation}
a_t^*\in\argmax_{a\in\mathcal A(x_t)}
\beta\log\!\left[
\frac{1}{|\mathcal T|}
\sum_{\tau\in\mathcal T}
\exp\!\left(\frac{\Delta_\tau(a)}{\beta}\right)
\right].
\label{eq:unified_plan}
\end{equation}
As $\beta\to0^+$, the objective becomes $\max_{\tau\in\mathcal T}\Delta_\tau(a)$. The planner greedily picks the action and horizon with the largest improvement, so each decision comes with an explicit horizon selection.

As $\beta\to\infty$, the objective becomes $|\mathcal T|^{-1}\sum_{\tau\in\mathcal T}\Delta_\tau(a)$. Unlike the greedy rule, which uses only the best horizon, averaging lets gains and losses at different horizons offset each other. The current-state term cancels, so candidates are ranked by $\sum_{\tau\in\mathcal T}e^{G_\theta(\hat x^a_{t+1},y_g,\tau)}$, a single potential over states for a fixed goal. With exact predictions and positive progress at every step, this potential increases along the trajectory, so no state is revisited.

We refer to these two limits as \emph{greedy planning} (GP) and
\emph{potential-ascent planning} (PAP), respectively, throughout
the paper.

\noindent\textbf{Closed-loop execution.}
At each step, it predicts the outcome of every candidate action, scores these outcomes by \cref{eq:unified_plan}, and executes $a_t^*$. The agent then observes the actual $x_{t+1}$ and repeats, until it reaches the goal or exhausts its budget. Because each prediction starts from an actual observation, errors do not compound over a multi-step rollout, while $G_\theta$ provides long-range guidance. Both $\theta$ and $\phi$ remain fixed throughout execution.

\Cref{app:algorithms} summarizes learning $\theta$, learning $\phi$, and closed-loop planning.

\section{Theoretical Understanding}
\label{sec:theory}
\label{sec:random_walk_theory}
\begin{figure}[!t]
\centering\includegraphics[width=\textwidth]{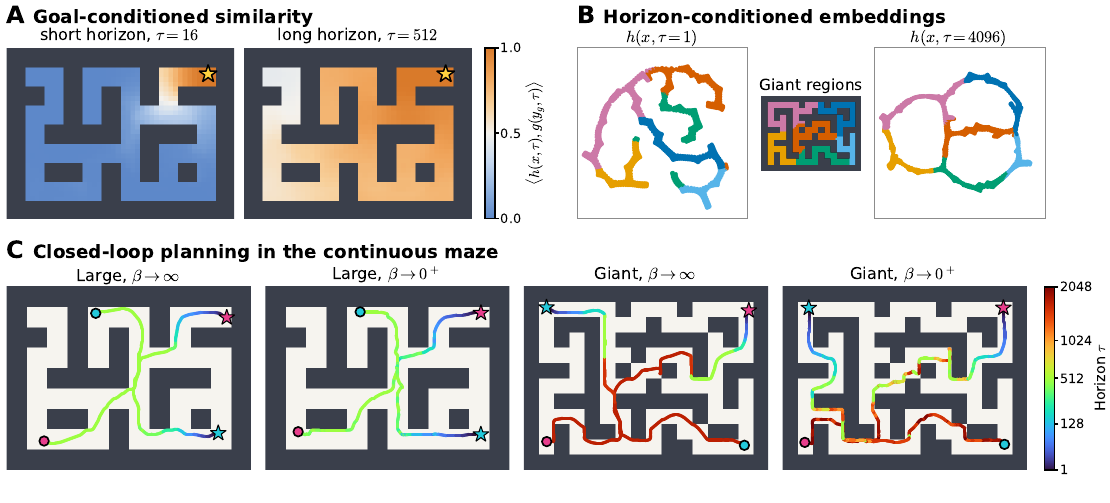}
\caption{
Emergent cognitive maps from random exploration.
\textbf{(A)} Score $G_\theta(x,y_g,\tau)$ over positions $x$ in Large,
with goal $y_g$ fixed (star). Short horizons ($\tau=16$) peak near
the goal, while long horizons ($\tau=512$) spread through connecting
passages.
\textbf{(B)} t-SNE of $h_\theta(x,\tau)$ in Giant at $\tau=1$ and
$4096$. At $\tau=1$, embeddings preserve local geodesic geometry,
while at $\tau=4096$, they reflect connectivity between distant
regions. 3D views appear in \cref{fig:tsne_3d}.
\textbf{(C)} GP and PAP routes through Large and Giant. Colour denotes
the selected horizon for GP and the dominant score-contributing
horizon for PAP. Matching markers pair starts ($\circ$) and
goals ($\star$).
}
\label{fig:overview}
\end{figure}

We study what temporal relations learned from random exploration reveal about an environment. The results concern the calibrated score $G^*(x,y,\tau)=\log[p_{\rm data}(y|x,\tau)/p_0(y)]$, the ideal log-density ratio our model approximates, and characterize the environmental structure it contains. \Cref{app:calibration} relates this score to the learned model.

\noindent\textbf{From random walks to the heat equation.}
Consider a symmetric local random walk over the free space, where a move into a wall leaves the agent in place. In free space, its steps are independent with zero mean and covariance $2\alpha I_n$. The constant $\alpha>0$ is set by the one-step transition kernel and does not depend on position or horizon. After $\tau$ steps, the root-mean-square displacement is $\sqrt{2n\alpha\tau}$, so doubling this diffusion scale takes four times as many steps. For regular domains and lattice approximations satisfying a reflected-Brownian invariance principle \citep{burdzy2008discrete}, we study the limit under diffusive rescaling, with time per step proportional to squared step length. Writing $\tau$ also for this rescaled time, with diffusion coefficient $\alpha$, its transition density satisfies $\partial_\tau p_{\rm data}=\alpha\,\Delta_y p_{\rm data}$.

\begin{proposition}[Short horizons encode geodesic geometry]
\label{prop:geodesic}
\label{prop:varadhan}
Consider the reflected Brownian motion above on a smooth, bounded, connected domain, with uniform $p_0$. Let $d_{\rm geo}(x,y)$ be the geodesic distance, the infimum of feasible path lengths between $x$ and $y$. For fixed interior $x,y$,
\begin{equation}
\lim_{\tau\downarrow0}\,-4\alpha\tau\,G^*(x,y,\tau)=d_{\rm geo}(x,y)^2.
\label{eq:varadhan}
\end{equation}
\end{proposition}
\begin{proof}[Proof sketch]
Varadhan's formula gives $-4\alpha\tau\log p_{\rm data}(y|x,\tau)\to d_{\rm geo}(x,y)^2$ \citep{varadhan1967heat,norris1997heat}, and $4\alpha\tau\log p_0(y)\to0$.
\end{proof}
This limit concerns rescaled diffusion time, rather than integer horizons on a fixed lattice (\cref{app:varadhan}). For the limiting diffusion, targets with shorter geodesic distances have larger scores at sufficiently short horizons, whatever the Euclidean distances. A nearby point behind a wall may require a long detour. At short horizons it then scores below a point that is farther in Euclidean distance but reachable along a shorter open route.

We now return to the discrete walk on a finite connected state space, with transition matrix $P$ and stationary distribution $\pi$, which is uniform because $P$ is symmetric.
\begin{proposition}[Long horizons encode multiscale connectivity]
\label{prop:spectral}
\label{prop:slow_modes}
Suppose $P$ is irreducible, aperiodic, and reversible, with eigenvalues $1=\lambda_1\ge\lambda_2\ge\cdots$ and eigenfunctions $\psi_j$ orthonormal under $\pi$. Set $p_0=\pi$. For every integer $\tau\ge1$ and every pair with $P^\tau(x,y)>0$,
\begin{equation}
G^*(x,y,\tau)=\log\Big[1+\sum_{j\ge2}\lambda_j^\tau\,\psi_j(x)\psi_j(y)\Big].
\label{eq:score_spectrum}
\end{equation}
\end{proposition}
\begin{proof}[Proof sketch]
Divide $P^\tau(x,y)=\pi(y)\sum_j\lambda_j^\tau\psi_j(x)\psi_j(y)$ by $\pi(y)$, separate $\psi_1=1$, and take the logarithm \citep{aldous2002reversible,coifman2006diffusion}.
\end{proof}
Terms with smaller $|\lambda_j|$ decay faster. Modes near $+1$ preserve differences between regions that exchange probability slowly. Picture two rooms joined by a narrow doorway. When movement between the rooms is slow relative to mixing within each room, a walk forgets which corner it started in long before it forgets which room it started in. Over these horizons, $G^*$ distinguishes the rooms more strongly than positions within each room (\cref{app:slow_modes}).

\noindent\textbf{The useful scale changes during planning.}
Choosing a horizon is like choosing the zoom level of a map. Zoomed in, the map shows the next turn but may leave the connecting passage out of view. Zoomed out, the passage appears, but the turns leading to it blur together. A navigator zooms out when the destination is far and zooms in as it approaches, and the horizon plays the same role. Progress compares the true goal probability from each candidate outcome with that from the current state. At a horizon too short for the walk to reach the goal, these probabilities are all nearly zero, so every candidate makes almost no progress. As the horizon goes to infinity, they all approach the stationary value $\pi(y_g)$, and progress again vanishes. Only intermediate horizons separate good moves from bad ones. In free space, the best horizon for a small step is roughly the one at which the walk's typical displacement reaches the goal, so it grows with the squared goal distance (\cref{app:spatial_scale}). In a maze, it also depends on the time needed to cross connecting passages. Empirically, the greedy planner selects shorter horizons when approaching the goal (\cref{fig:overview}C).

\section{Experiments}
\label{sec:experiments}
\providecommand{\pmc}[2]{#1\,{\scriptsize\textcolor{black!55}{$\pm$#2}}}   %
\providecommand{\fade}[1]{\textcolor{black!55}{#1}}                            %
\definecolor{oursbg}{gray}{0.92}                                              %

We evaluate goal reaching, horizon selection, data efficiency, first-person observations, and manipulation. \Cref{sec:experiment_setup} specifies the data and evaluation conditions and \cref{sec:experiment_results} reports the results.

\subsection{Experimental Setup}
\label{sec:experiment_setup}
\label{sec:experimental_setup}

\noindent\textbf{Datasets and Environments.}
We study three settings.
\emph{(1) Maze navigation:} OGBench PointMaze Large and Giant \citep{park2025ogbench}, with position or $64\!\times\!64$ overhead images. Our main models use 4M lattice transitions: eight neighbouring moves or stay sampled uniformly on a 1-unit grid, with blocked moves replaced by stay and five simulator steps per transition. Reproduced baselines use 1M transitions from uniformly sampled actions (20 episodes of 50k steps). Our local dynamics uses 1M transitions on Large and 10M on Giant. \textsc{Navigate} uses OGBench expert data.
\emph{(2) Egocentric navigation:} a first-person Large maze and Habitat's Van Gogh Room and Apartment~1 \citep{savva2019habitat}. Maze inputs are images with optional position and/or heading. Habitat models use one 8M-step walk per scene; lattice planning compares position alone with images and position, and continuous planning adds heading.
\emph{(3) Manipulation:} OGBench cube-single with expert-only and mixed data. The expert-only setting follows LeWM's data and evaluation protocol \citep{maes2026lewm} and uses its reported baseline results. The mixed setting combines expert trajectories with noisy oracle trajectories, generated by adding temporally correlated noise to the oracle's waypoint plan and replacing each action with a random one with probability $0.1$.

\noindent\textbf{Evaluation.}
We report success rate (SR) and success weighted by path length (SPL) \citep{anderson2018evaluation}:
\begin{equation}
\mathrm{SR}=\frac{1}{M}\sum_{i=1}^{M}S_i,
\qquad
\mathrm{SPL}=\frac{1}{M}\sum_{i=1}^{M}S_i\frac{\ell_i}{\max(p_i,\ell_i)},
\label{eq:spl}
\end{equation}
Here $M$ counts episodes, $S_i$ indicates success, $p_i$ is executed path length, and $\ell_i$ is shortest-path length to the goal region.
Setting~(1) uses five official tasks (20 episodes each, 1k-simulator-step budget) and 20 fixed start--goal pairs per maze (geodesic separation $\ge20$ cells, 5k-simulator-step budget). Success requires reaching within one unit of the goal. Candidate outcomes come from learned local dynamics, without simulator queries before action execution.
Setting~(2) reuses the maze tasks. Habitat evaluates its two most distant reachable corner pairs in both directions: 25 episodes per task, random initial headings, and a 5k-lattice-action budget. Success requires reaching within one lattice move of the goal or within $0.3$\,m in continuous evaluation.
In setting~(3), the goal is the observation 25 steps after the initial state in the same dataset trajectory. Each episode allows 50 steps and succeeds if the cube center comes within $0.04$~m of its goal position at any step. We report SR only. Mixed-data comparisons with LeWM use 1{,}000 episodes from the held-out split.

\noindent\textbf{Comparisons.}
\emph{World models} (LeWM, DINO-WM) plan through learned dynamics using their own planners. \emph{Offline goal-conditioned RL} (GCIQL, HIQL, HILP, QRL) executes policies learned from offline data. Random actions and a shortest-path oracle provide reference controls.

\subsection{Results}
\label{sec:experiment_results}
\label{sec:experimental_results}
\begin{figure}[t]
\centering
\includegraphics[width=0.9\linewidth]{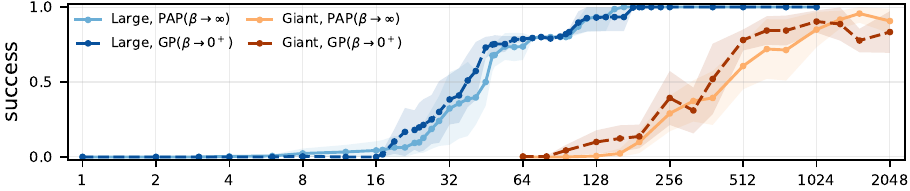}
\caption{Official-task success versus maximum planning horizon. Color distinguishes Large and Giant and line style distinguishes GP and PAP. Lines show three-seed means and bands one standard deviation. Planning uses every integer horizon from $1$ to $\tau_{\max}$.}
\vspace{-1.2em}
\label{fig:menu}
\end{figure}

\begin{figure}[t]
\centering
\includegraphics[width=\textwidth]{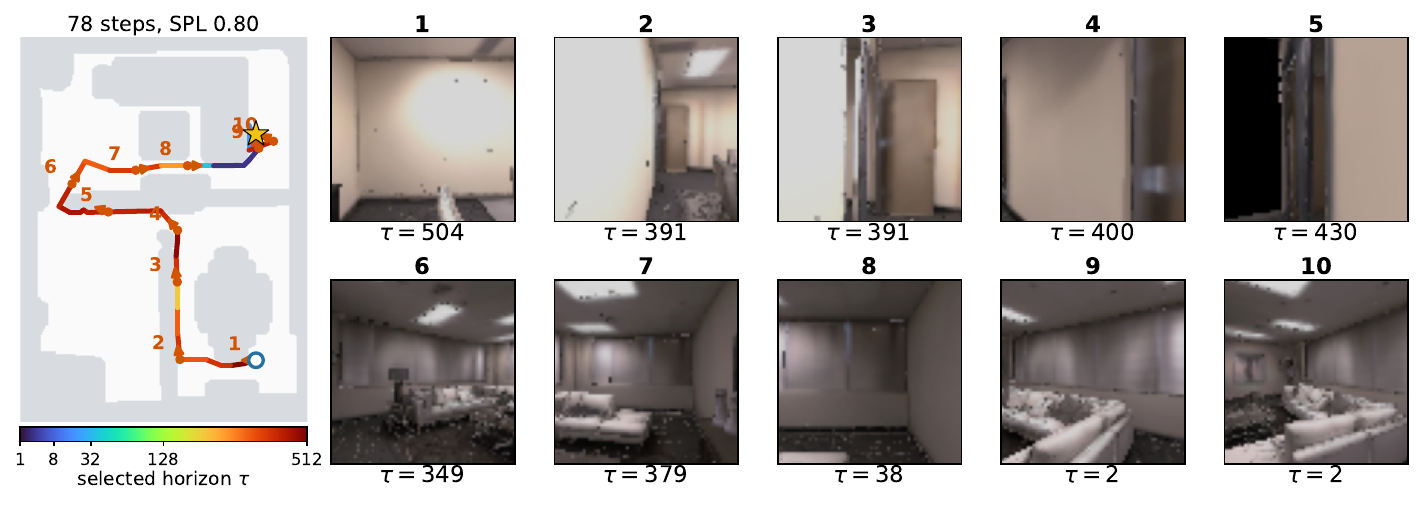}
\caption{An egocentric plan in Habitat Apartment~1 by the model that plans from the image and the position (horizons capped at 512, no constraint on consecutive headings). Left: the route, coloured by the selected horizon, from the start ($\circ$) to the goal ($\star$). Right: the agent's view at ten decisions and the horizon selected there.}
\label{fig:first_person}
\vspace{-1em}
\end{figure}

We first show that random exploration supports long-range planning from both states and images, and that the learned representation acquires multiscale spatial structure consistent with the random-walk analysis (\cref{tab:maze}, \cref{fig:overview}). We then study how the available horizons and exploration coverage shape planning performance (\cref{fig:menu}, \cref{tab:data_efficiency}). Finally, we test planning from egocentric observations and manipulation with suboptimal data (\cref{fig:first_person}, \cref{tab:manipulation}).

\begin{table}[t]
\centering\scriptsize
\caption{PointMaze goal reaching from states (A) and images (B) on official tasks and random start--goal pairs. The GCIQL, QRL, and HIQL \textsc{Navigate} results are quoted from \citet{park2025ogbench}.}
\label{tab:maze}
\textbf{A\quad State}\par\smallskip
\setlength{\tabcolsep}{2pt}
\renewcommand{\arraystretch}{0.9}
\begin{tabularx}{\linewidth}{@{}l*{10}{C}@{}}
\toprule
 & \multicolumn{6}{c}{\textbf{Official tasks}} & \multicolumn{4}{c}{\textbf{Random pairs}}\\
\cmidrule(lr){2-7}\cmidrule(lr){8-11}
 & \multicolumn{2}{c}{\textsc{Navigate}} & \multicolumn{4}{c}{Random Data} & \multicolumn{4}{c}{Random Data}\\
\cmidrule(lr){2-3}\cmidrule(lr){4-7}\cmidrule(lr){8-11}
 & Large & Giant & \multicolumn{2}{c}{Large} & \multicolumn{2}{c}{Giant} & \multicolumn{2}{c}{Large} & \multicolumn{2}{c}{Giant}\\
\cmidrule(lr){2-2}\cmidrule(lr){3-3}\cmidrule(lr){4-5}\cmidrule(lr){6-7} \cmidrule(lr){8-9}\cmidrule(lr){10-11}
\textbf{Method} & SR$\uparrow$ & SR$\uparrow$ & SR$\uparrow$ & SPL$\uparrow$ & SR$\uparrow$ & SPL$\uparrow$ & SR$\uparrow$ & SPL$\uparrow$ & SR$\uparrow$ & SPL$\uparrow$\\
\midrule
\fade{Random policy (floor)} & \fade{0.00} & \fade{0.00} & \fade{0.00} & \fade{0.00} & \fade{0.00} & \fade{0.00} & \fade{0.00} & \fade{0.00} & \fade{0.00} & \fade{0.00}\\
\fade{Shortest-path oracle} & \fade{1.00} & \fade{1.00} & \fade{1.00} & \fade{0.94} & \fade{1.00} & \fade{0.98} & \fade{1.00} & \fade{0.95} & \fade{1.00} & \fade{0.95}\\
\midrule
GCIQL \citep{kostrikov2022iql} & 0.34 & 0.00 & 0.00 & 0.00 & 0.00 & 0.00 & 0.15 & 0.15 & 0.07 & 0.06\\
QRL \citep{wang2023qrl} & 0.86 & 0.68 & 0.33 & 0.28 & 0.02 & 0.01 & 0.40 & 0.34 & 0.30 & 0.25\\
HIQL \citep{park2023hiql} & 0.58 & 0.46 & 0.00 & 0.00 & 0.00 & 0.00 & 0.12 & 0.11 & 0.10 & 0.09\\
HILP \citep{park2024hilp} & --- & --- & 0.49 & 0.47 & 0.00 & 0.00 & 0.47 & 0.45 & 0.18 & 0.18\\
\midrule
LeWM \citep{maes2026lewm} & 0.08 & 0.00 & 0.10 & 0.03 & 0.00 & 0.00 & 0.45 & 0.14 & 0.10 & 0.04\\
\midrule
\rowcolor{oursbg} PAP ($\beta\to\infty$) & --- & --- & 0.75 & 0.66 & \textbf{0.79} & \textbf{0.64} & 0.92 & 0.83 & \textbf{0.85} & \textbf{0.77}\\
\rowcolor{oursbg} GP ($\beta\to0^+$) & --- & --- & \textbf{0.92} & \textbf{0.83} & 0.76 & 0.58 & \textbf{0.97} & \textbf{0.88} & \textbf{0.85} & 0.75\\
\bottomrule
\end{tabularx}
\par\vspace{4pt}
\textbf{B\quad Vision}\par\smallskip
\setlength{\tabcolsep}{3pt}
\renewcommand{\arraystretch}{0.9}
\begin{tabularx}{\linewidth}{@{}l*{6}{C}@{}}
\toprule
 & \multicolumn{2}{c}{\textbf{Official tasks}} & \multicolumn{4}{c}{\textbf{Random pairs}}\\
\cmidrule(lr){2-3}\cmidrule(lr){4-7}
 & Large & Giant & \multicolumn{2}{c}{Large} & \multicolumn{2}{c}{Giant}\\
\cmidrule(lr){2-2}\cmidrule(lr){3-3}\cmidrule(lr){4-5}\cmidrule(lr){6-7}
\textbf{Method} & SR$\uparrow$ & SR$\uparrow$ & SR$\uparrow$ & SPL$\uparrow$ & SR$\uparrow$ & SPL$\uparrow$\\
\midrule
LeWM \citep{maes2026lewm} & 0.00 & 0.00 & 0.00 & 0.00 & 0.00 & 0.00\\
DINO-WM \citep{zhou2025dinowm} & 0.03 & 0.00 & 0.35 & 0.23 & 0.00 & 0.00\\
\midrule
\rowcolor{oursbg} PAP ($\beta\to\infty$) & 0.79 & \textbf{0.53} & \textbf{0.95} & 0.83 & 0.70 & 0.63\\
\rowcolor{oursbg} GP ($\beta\to0^+$) & \textbf{0.91} & 0.39 & \textbf{0.95} & \textbf{0.85} & \textbf{0.78} & \textbf{0.67}\\
\bottomrule
\end{tabularx}
\vspace{-1.5em}
\end{table}

\noindent\textbf{Planning from random exploration.}
\label{sec:maze_results}
\label{sec:continuous}
Temporal relations learned from random exploration support long-range goal reaching from both states and images (\cref{tab:maze}). Averaged over three training seeds, GP and PAP reach $0.97$ and $0.92$ of the random start--goal pairs on Large and $0.85$ of them on Giant from states. From images, GP and PAP reach $0.95$ of the random pairs on Large with SPL $0.85$ and $0.83$, and $0.78$ and $0.70$ of them on Giant. Most reported baselines achieve little or no success on the official tasks in the random-data setting. Unlike goal-conditioned RL, our model learns temporal relations under the exploration behavior without policy-improvement training. These relations provide long-range guidance for local action selection under both planning objectives.

\noindent\textbf{Emergent cognitive maps.}
The learned temporal representation exhibits properties of a multiscale cognitive map \citep{tolman1948cognitive} in its scores, embeddings, and planned routes. For a fixed goal, short-horizon embedding similarities $\langle h_\theta(x,\tau),g_\theta(y_g,\tau)\rangle$ emphasize nearby locations, while longer horizons reveal connections through passages (\cref{fig:overview}A). The embeddings $h_\theta(x,\tau)$ represent the same environment at different spatial scales (\cref{fig:overview}B). Distance probes quantify this change: shorter horizons better preserve local geodesic geometry, while longer horizons better reflect connectivity between distant states (\cref{tab:latent_rank}). These patterns are consistent with the random-walk analysis in \cref{sec:theory} and emerge without distance or route supervision. In planning, GP and PAP use this structure to turn random exploration into efficient, goal-directed routes through the maze (\cref{fig:overview}C).

\noindent\textbf{Horizon selection and aggregation.}
\label{sec:horizon}
The larger maze benefits from longer planning horizons (\cref{fig:menu}): Large reaches about $0.8$ success at $\tau_{\max}=64$, whereas Giant improves mainly as the range grows from $256$ to about $1{,}024$. Further extension does not consistently improve performance. GP selects the largest progress at any horizon, while PAP aggregates progress across horizons. The two achieve comparable success on random maze pairs (\cref{tab:maze}), while PAP achieves higher success in the cube-single comparison ($0.92$ versus $0.84$; \cref{tab:manipulation}). Along the routes in \cref{fig:overview}C, GP's selected horizon and PAP's dominant score-contributing horizon generally shorten near the goal, although PAP continues to aggregate all horizons. This change occurs without a prescribed schedule and is consistent with the spatial-scale analysis in \cref{sec:theory}.

\begin{table}[t]
\centering\scriptsize
\begin{minipage}[b]{0.51\linewidth}
\centering
\caption{Goal reaching vs.\ walk length and exploration coverage $c$. Full coverage ($c=1$) repeats the 4M reference.}
\label{tab:data_efficiency}
\setlength{\tabcolsep}{2pt}
\begin{tabularx}{\linewidth}{@{}l*{4}{C}@{}}
\toprule
 & \multicolumn{2}{c}{GP} & \multicolumn{2}{c}{PAP}\\
\cmidrule(lr){2-3}\cmidrule(lr){4-5}
Walk (steps) & Large & Giant & Large & Giant\\
\midrule
0.1M & \pmc{0.45}{0.19} & 0.01 & \pmc{0.26}{0.18} & 0.00\\
0.4M & \pmc{0.74}{0.06} & \pmc{0.15}{0.12} & \pmc{0.68}{0.16} & \pmc{0.13}{0.09}\\
1M   & \pmc{0.91}{0.09} & \pmc{0.49}{0.09} & \pmc{0.68}{0.18} & \pmc{0.50}{0.21}\\
\rowcolor{oursbg} 4M & \pmc{0.92}{0.09} & \pmc{0.76}{0.08} & \pmc{0.75}{0.13} & \pmc{0.79}{0.14}\\
\midrule
$c=0.25$ & \pmc{0.09}{0.09} & \pmc{0.00}{0.00} & \pmc{0.27}{0.09} & \pmc{0.19}{0.16}\\
$c=0.50$ & \pmc{0.24}{0.17} & 0.03 & \pmc{0.26}{0.10} & 0.02\\
$c=0.75$ & \pmc{0.56}{0.14} & 0.36 & \pmc{0.59}{0.22} & 0.20\\
\rowcolor{oursbg} $c=1$ & \pmc{0.92}{0.09} & \pmc{0.76}{0.08} & \pmc{0.75}{0.13} & \pmc{0.79}{0.14}\\
\bottomrule
\end{tabularx}
\end{minipage}\hfill
\begin{minipage}[b]{0.45\linewidth}
\centering
\caption{Cube-single: mixed data, held-out episodes (top); expert data, LeWM's protocol and baselines (bottom).}
\label{tab:manipulation}
\setlength{\tabcolsep}{3pt}
\renewcommand{\arraystretch}{1.06}
\begin{tabular}{@{}lcc@{}}
\toprule
Expert data ratio& PAP & LeWM\\
\midrule
0\%   & \pmc{0.908}{0.009} & \pmc{0.633}{0.013}\\
25\%  & \pmc{\textbf{0.933}}{0.008} & \pmc{0.626}{0.009}\\
50\%  & \pmc{0.926}{0.006} & \pmc{0.623}{0.015}\\
75\%  & \pmc{0.905}{0.010} & \pmc{0.625}{0.010}\\
100\% & \pmc{0.876}{0.010} & \pmc{\textbf{0.710}}{0.012}\\
\midrule
PLDM \citep{sobal2025pldm} & \multicolumn{2}{c}{0.65}\\
LeWM \citep{maes2026lewm} & \multicolumn{2}{c}{0.74}\\
DINO-WM \citep{zhou2025dinowm} & \multicolumn{2}{c}{0.86}\\
 GP/PAP & \multicolumn{2}{c}{0.84/\textbf{0.92}}\\
\bottomrule
\end{tabular}
\end{minipage}
\vspace{-2em}
\end{table}

\noindent\textbf{Exploration coverage.}
\label{sec:data_efficiency}
Every walk in the length rows visits every free cell (a 0.1M-step walk already passes each Large cell about 136 times and each Giant cell about 73 times), so these rows vary the number of transitions per cell rather than coverage. The coverage fraction $c$ specifies the part of the maze available to a 4M-step walk. Spreads denote standard deviations. The 4M and $c=1$ rows are the models of \cref{tab:maze}. Every row plans with all horizons from 1 to 8192 through the learned local dynamics.
Both data volume and state coverage affect planning performance. On Large, increasing the explored fraction from $c=0.25$ to $c=0.75$ raises success from $0.09$ to $0.56$ for GP and from $0.27$ to $0.59$ for PAP, and full coverage reaches $0.92$ and $0.75$ (\cref{tab:data_efficiency}). With unrestricted exploration, GP on Large rises from $0.45$ after $0.1$M steps to $0.91$ after $1$M steps. Giant needs longer walks: GP reaches $0.01$ after $0.1$M steps, $0.49$ after $1$M and $0.76$ after $4$M. Planning therefore improves with both walk length and coverage, and the larger maze requires more exploration.

\noindent\textbf{Egocentric observations.}
\label{sec:first_person}
Planning extends to first-person observations. In Habitat, we train on one 8M-step walk per scene, collected with $30^\circ$ turns and $0.5$\,m forward or backward steps on a $0.25$\,m lattice. On Apartment~1's four corner tasks, images and position achieve $0.71\pm0.12$ success at $\tau_{\max}=8{,}192$, compared with $0.38$ for a random policy. The position-only planner repeatedly turns in place and reaches no goals. Under the continuous protocol (36 headings, walking $0.75$\,m to the scored candidate before re-planning, no constraint on consecutive headings), the image-and-position model reaches a far pair of the apartment from all three start headings with horizons capped at $512$ (SR $1.00$, SPL $0.77$; position alone: SR $1.00$, SPL $0.79$). The full horizon range strands the image model, since the longest horizons dominate at headings unseen in training, as on Giant beyond $\tau_{\max}\approx1{,}000$ (\cref{fig:menu}). Along a successful route, the selected horizon shortens as the goal comes into view (\cref{fig:first_person}). In the egocentric maze, we find position helps more than heading.

\noindent\textbf{Manipulation with suboptimal data.}
Our model supports effective manipulation planning from suboptimal trajectories. On cube-single under LeWM's evaluation protocol, PAP reaches success $0.92$ while GP reaches $0.84$.
We also vary the expert fraction at a fixed training budget and compare PAP with LeWM on matched data (\cref{tab:manipulation}). Our model achieves higher success at every mixture. Noisy-oracle data alone yields success $0.908$, exceeding $0.876$ with expert-only data, and the highest reported success is $0.933$ at $25\%$ expert data. Our temporal model learns how states are connected from the observed trajectories. Noise may expose a wider range of these connections, making suboptimal trajectories useful for planning even when they are less efficient at reaching goals.

\section{Related Work}
\label{sec:related}
\noindent\textbf{World models and temporal representations.}
Latent world models support goal reaching through action-sequence
optimization. DINO-WM predicts dynamics in pretrained visual features
\citep{zhou2025dinowm}, while Temporal Straightening jointly learns
representations and dynamics with a curvature regularizer
\citep{wang2026temporal}. Temporal representation learning provides
another source of structure. Contrastive predictive coding
discriminates future observations \citep{oord2018cpc}, and
Contrastive RL connects temporal associations to goal-conditioned
values \citep{eysenbach2022crl}. Successor and forward--backward
representations relate future occupancy to rewards
\citep{dayan1993sr,touati2021fb}, while contrastive representations
support planning through interpolation and temporal reasoning
\citep{eysenbach2024interpolation,ziarko2025crtr}.
Our model separates long-range guidance learned from observation pairs
from local action-outcome prediction.

\noindent\textbf{Random-walk geometry and planning.}
Diffusion maps connect transition statistics to geometry
\citep{coifman2006diffusion}. \citet{zhao2025placecells} model
place-cell populations through position embeddings fitted to
predefined random-walk kernels. We learn conditional
temporal log-density ratios from sampled observation pairs through
NCE, using horizon-conditioned state or image encoders. These scores
guide action selection through horizon-specific progress in GP or
a goal-dependent potential in PAP.
Quasimetric learning targets directed optimal goal distances
\citep{wang2023qrl}, while replay-buffer search composes local
connections \citep{eysenbach2019sorb}. Our temporal scores describe
exploration behavior, with geometric and connectivity interpretations
under the random-walk assumptions in \cref{sec:theory}.

\section{Limitations and Future Work}
\label{sec:discussion}

Random exploration can teach representations that support
goal-directed behavior without policy-improvement training.
Our current models learn within individual environments.
Planning in unseen environments raises the question of what
temporal structure transfers and what must be learned from
new experience. Behavioral timescale synaptic plasticity
supports rapid place-field formation~\citep{bittner2017behavioral}
and may inspire adaptation from limited experience.
Egocentric planning presents a related challenge: a single
view may not determine the state. Observation histories
could help recover spatial context and reduce reliance on
explicit position inputs. Scaling training to larger datasets
across diverse environments would let us study whether shared
temporal representations support transfer and rapid adaptation.

The connection to animal cognition also remains incomplete.
Although the embeddings are learned, the learning objective,
horizon conditioning, and planning rules are designed explicitly.
Our framework does not explain how animals acquire these mechanisms.
A broader account would need to explain how evolution shapes
general learning and control mechanisms through which both
representations and planning strategies emerge from experience.

\section*{Acknowledgments}
We thank Chenxin Tao for insightful discussions during the spring of 2025, and Minglu Zhao and Dehong Xu for earlier collaborations. Y.~W. is partially supported by NSF DMS-2415226, DARPA W912CG25CA007 and research gift funds from Amazon and Qualcomm. T.~G. is supported by NSF under Award No. 2610649 and by NERSC through DDR-ERCAP0035256.
\bibliography{references}
\bibliographystyle{iclr2027_conference}

\clearpage
\appendix
\addtocontents{toc}{\protect\setcounter{tocdepth}{2}}
\begingroup
\renewcommand{\contentsname}{Appendix}
\hypersetup{linktoc=all}
\tableofcontents
\endgroup
\clearpage
\section{Extended related work}
\paragraph{Latent world models.}
Latent world models predict how a latent state evolves under actions, rather than predicting future frames, and plan or learn policies in that latent state space. They can be grouped by the training signal that shapes the latent state.
Reconstruction-based models learn the latent through a decoder that reconstructs observations~\citep{ha2018worldmodels,hafner2019planet,hafner2020dreamer,hafner2025dreamerv3,micheli2023iris}. Reconstruction provides a dense training signal, but it pushes the latent to encode visual detail regardless of its relevance to control.
Reward-driven models remove the decoder and shape the latent through task signals such as reward and value prediction, often combined with a latent consistency loss~\citep{schrittwieser2020muzero,hansen2022tdmpc,hansen2024tdmpc2}. These objectives encourage the latent to retain information useful for reward prediction and control, but require reward labels and tie representation learning to the training tasks.
JEPA-style world models need neither a decoder nor rewards: they are trained to predict future embeddings from reward-free trajectories, and they plan by reaching a goal embedding.
With no reconstruction or reward signal to anchor the latent, their central challenge is representation collapse. One line of work avoids it by freezing a pretrained visual encoder during action-conditioned dynamics training, as in DINO-WM and V-JEPA~2-AC~\citep{zhou2025dinowm,assran2025vjepa2}, while another trains the encoder end-to-end with explicit anti-collapse regularization~\citep{sobal2025pldm,maes2026lewm}.
LeWorldModel uses a single Gaussian regularizer~\citep{maes2026lewm}. Subsequent work studies alternative regularization~\citep{yu2026qqworld}, dynamics models~\citep{huo2026flowjepa,gao2026fastlewm}, auxiliary physical supervision~\citep{liu2026grounded}, hierarchical planning~\citep{caselli2026mindthegap}, and action optimization with a terminal state-space energy over a frozen world model~\citep{pham2026leap}.

\paragraph{Energy-based models and inference for planning.}
Energy-based models (EBMs) learn distributions through energy functions, with foundational work connecting generative ConvNets to analysis-by-synthesis learning and cooperative training with latent generators~\citep{xie2016generative,xie2018cooperative}. Noise-contrastive estimation learns density ratios from data and reference samples~\citep{gutmann2010nce,yu2026noisier}. For control, learned energies serve as costs in inverse optimal control~\citep{xu2023inverse}, guide state-sequence planning~\citep{du2019model}, and regularize planning toward likely transitions~\citep{boney2019regularizing}. Energy-based motion planners and diffusion planners further support trajectory optimization and compositional conditioning~\citep{urain2022learning,urain2023se3,luo2024potential,gkanatsios2023energy,yang2023compositional,janner2022planning,ajay2023conditional}. A related line moves inference into latent spaces, using latent plans for return-conditioned generation, online adaptation, and policy improvement~\citep{kong2024latent,noh2025latent,qin2025generative}, or latent EBMs for task adaptation and continuous-time dynamics~\citep{boebm,cheng2024latent}. Beyond control, latent inference supports molecular design~\citep{kong2023molecule,kong2024molecule}, knowledge-grounded dialogue~\citep{xu2023diverse}, and reasoning through latent thoughts and iterative refinement~\citep{kong2025thought,kong2026rethinking}. FaST studies a complementary form of adaptive computation, switching between fast responses and slower inference for visual reasoning~\citep{sun2025visual}. Spatial representation models connect navigation to head-direction coding~\citep{zhao2025head}, grid-cell embeddings that preserve local distances up to a scale factor~\citep{xu2025gridcells}, and place-cell embeddings of multiscale random-walk kernels~\citep{zhao2025placecells}. Our model learns conditional temporal log-density ratios from observation pairs through NCE. These scores provide long-range goal guidance for local action evaluation. GP retains the joint move-and-horizon maximization of \citet{zhao2025placecells}, applied to estimated density-ratio progress, while our temperature objective also yields PAP, which aggregates progress across horizons into a single goal-dependent potential.

\newpage
\section{Derivations and Proofs}
\label{app:proofs}
We first show what NCE learns from temporal pairs, then derive the geometric information in these relations at short and long horizons. We next examine how the embedding model represents this information and how GP and PAP use it to evaluate progress toward a goal.

The geometric results concern the ideal score $G^*=\log[p_{\rm data}(y|x,\tau)/p_0(y)]$ under the stated diffusion or Markov assumptions. For the Markov analysis to apply directly to observations, each observation must determine the state. These results describe the ideal temporal relations; learning at finitely many horizons does not by itself establish the same limiting behavior for $G_\theta$. The planning identities, however, hold for any finite learned scores.

\subsection{Density-ratio learning and normalization}
\label{app:calibration}
NCE learns temporal association by distinguishing observed pairs from pairs with independently sampled targets. Assume that the reference $p_0$ covers the conditional target support. At a sampled horizon, the two pair densities are $p_{\rm data}(x,y|\tau)$ and $p_{\rm data}(x|\tau)p_0(y)$. With $N$ negatives per positive, the class prior odds are $1:N$. The source marginal cancels from the posterior odds, leaving $p_{\rm data}(y|x,\tau)/(Np_0(y))$.

The classifier $\sigma(G-\log N)$ has odds $e^G/N$. Matching these odds gives the unrestricted population minimizer of the logistic loss in \cref{eq:nce}:
\[
G^*(x,y,\tau)=\log\frac{p_{\rm data}(y|x,\tau)}{p_0(y)}.
\]
This equality holds on sampled sources and horizons where both densities are positive \citep{gutmann2010nce,ma2018conditionalnce}. Where $p_0(y)>0$ and the data conditional is zero, the optimum is approached as $G\to-\infty$, and we interpret $e^{G^*}=0$.

The fitted score need not satisfy the same normalization as this ideal ratio. For the normalized EBM in \cref{eq:conditional_model}, the exact relationship is
\begin{equation}
\log\frac{p_\theta(y|x,\tau)}{p_0(y)}
=G_\theta(x,y,\tau)-\log\E_{p_0(y')}\!\left[e^{G_\theta(x,y',\tau)}\right].
\label{eq:normalization_gap}
\end{equation}
The expectation equals $e^{\beta_1(\tau)}Z_\theta(x,\tau)$ and measures how far the fitted ratio is from integrating to one. The ideal ratio integrates to one at every source. A shared offset $\beta_1(\tau)$ cannot enforce this when $Z_\theta(x,\tau)$ varies with $x$. This matters during planning: candidates are different sources paired with the same goal, so a source-dependent correction can change their ranking. The planner uses the fitted score $G_\theta$.

\subsection{Short horizons: geodesic geometry and spatial scale}
\label{app:varadhan}
Short temporal relations reveal path geometry through the diffusion limit of the walk. To obtain this limit, shrink free-space steps by $\delta$ and assign each step time $\delta^2$. The covariance per unit time remains $2\alpha I_n$. Under the invariance principle assumed in \cref{sec:theory}, the walk converges to reflected Brownian motion with generator $\alpha\Delta$ \citep{burdzy2008discrete}. Reflection prevents probability from flowing through walls. We first take this continuum limit, then let diffusion time tend to zero.

\begin{proof}[Proof of \Cref{prop:varadhan}]
On the smooth, bounded, connected domain of the proposition, Varadhan's formula for the reflecting heat kernel gives, for fixed interior $x,y$,
\[
-4\alpha\tau G^*(x,y,\tau)
=-4\alpha\tau\log p_{\rm data}(y|x,\tau)+4\alpha\tau\log p_0(y)
\longrightarrow d_{\rm geo}(x,y)^2,
\]
because the uniform reference is positive and independent of $\tau$ \citep{varadhan1967heat,norris1997heat}.
\end{proof}
This limit also explains how the score ranks candidate moves. Among a finite set of interior candidates, suppose one has strictly shorter geodesic distance to $y_g$ than all the others. It maximizes $G^*(\cdot,y_g,\tau)$ for sufficiently small $\tau$. Indeed, its score advantage over each competitor, multiplied by $4\alpha\tau$, tends to the strictly positive difference of squared distances.

The order of limits matters. On a fixed graph, let $\ell\ge1$ be the smallest number of edges between distinct reachable states $x,y$. Then $P^k(x,y)=0$ for $k<\ell$. Even the continuous-time walk on this graph satisfies
\[
[e^{\tau(P-I)}](x,y)=\tau^\ell P^\ell(x,y)/\ell!+O(\tau^{\ell+1}).
\]
Its logarithm therefore scales as $\ell\log\tau$, rather than $-d_{\rm geo}^2/(4\alpha\tau)$ \citep{steinerberger2018varadhan}.

\noindent\textbf{The horizon giving the strongest local progress.}
\label{app:spatial_scale}
The geodesic limit describes which candidate scores higher. To see which horizon gives the strongest progress signal, consider diffusion in free space $\mathbb R^n$, whose transition density is
\begin{equation}
p_{\rm data}(y|x,\tau)
=(4\pi\alpha\tau)^{-n/2}\exp\!\left(-\frac{\|y-x\|_2^2}{4\alpha\tau}\right).
\label{eq:free_space_heat_kernel}
\end{equation}
Fix $x\ne y_g$, and write the goal distance as $r=\|y_g-x\|_2$ and the direction toward it as $u=(y_g-x)/r$. For a fixed reference with $0<p_0(y_g)<\infty$, the change in the ratio per unit step toward the goal is
\begin{equation}
\left.\frac{\partial}{\partial\epsilon}e^{G^*(x+\epsilon u,y_g,\tau)}\right|_{\epsilon=0}
=\frac{r}{2\alpha\tau}\frac{p_{\rm data}(y_g|x,\tau)}{p_0(y_g)}.
\label{eq:free_space_progress}
\end{equation}
We maximize this local progress over the horizon. Its logarithmic derivative in $\tau$ is $-(n+2)/(2\tau)+r^2/(4\alpha\tau^2)$, which changes from positive to negative at
\begin{equation}
\tau^*=\frac{r^2}{2(n+2)\alpha},
\qquad
\sqrt{2n\alpha\tau^*}=r\sqrt{\frac{n}{n+2}}.
\label{eq:free_space_optimal_horizon}
\end{equation}
The strongest local signal therefore occurs when the walk's root-mean-square displacement is a fixed fraction of the goal distance. At shorter horizons, little probability has reached the goal; at longer horizons, probability has spread over a larger region. Their balance gives the squared-distance scaling in \cref{sec:theory}. This calculation concerns infinitesimal motion in free space; finite moves, walls, and passages can change the maximizing horizon.

\subsection{Long horizons: connectivity and mixing}
\label{app:slow_modes}
At longer horizons, the walk loses information about its starting position. The spectral expansion describes which differences persist as this happens. We now return to a finite state space $\mathcal S$ and a discrete walk with irreducible, aperiodic, reversible transition matrix $P$. Let $\pi$ be its stationary distribution, set $p_0=\pi$, and order the eigenvalues as $1=\lambda_1>\lambda_2\ge\cdots$. At integer horizons, $p_{\rm data}(y|x,\tau)=P^\tau(x,y)$.

\begin{proof}[Proof of \Cref{prop:slow_modes}]
Put $D_\pi=\operatorname{diag}(\pi)$. Detailed balance makes $S=D_\pi^{1/2}PD_\pi^{-1/2}$ symmetric. If $S\phi_j=\lambda_j\phi_j$ with orthonormal $\phi_j$, then $\psi_j=D_\pi^{-1/2}\phi_j$ are orthonormal under $\pi$. Irreducibility gives $\lambda_1=1$, $\psi_1=1$; aperiodicity gives $|\lambda_j|<1$ for $j\ge2$. Expanding $P^\tau=D_\pi^{-1/2}S^\tau D_\pi^{1/2}$ yields
\begin{equation}
e^{G^*(x,y,\tau)}=\frac{P^\tau(x,y)}{\pi(y)}
=1+\sum_{j\ge2}\lambda_j^\tau\psi_j(x)\psi_j(y).
\label{eq:spectral_ratio}
\end{equation}
Taking logarithms where $P^\tau(x,y)>0$ proves the proposition \citep{aldous2002reversible,coifman2006diffusion}; the ratio identity also holds at zero transition probability.
\end{proof}

\noindent\textbf{Why weak connections produce slow modes.}
To connect the spectrum to the environment, we measure how much a function changes across transitions of the walk. Write $\langle f,g\rangle_\pi=\sum_x\pi(x)f(x)g(x)$. Expanding the squared change and using stationarity gives
\begin{equation}
\begin{aligned}
\mathcal E(f)&:=\frac12\sum_{u,v}\pi(u)P(u,v)[f(u)-f(v)]^2
=\langle f,(I-P)f\rangle_\pi,\\
\mathcal E(\psi_j)&=1-\lambda_j.
\end{aligned}
\label{eq:dirichlet}
\end{equation}
An eigenfunction with eigenvalue near one therefore varies little across transitions, on average. Now divide the states into regions $A$ and $A^c$, with $0<\pi(A)<1$. The function $f=(\mathbf1_A-\pi(A))/\sqrt{\pi(A)\pi(A^c)}$ is constant within each region and has zero mean and unit norm under $\pi$. Substituting it into the variational formula
$1-\lambda_2=\min_{\langle f,1\rangle_\pi=0,\,\langle f,f\rangle_\pi=1}\mathcal E(f)$ gives
\begin{equation}
1-\lambda_2\le\frac{Q(A,A^c)}{\pi(A)\pi(A^c)},
\qquad Q(A,A^c)=\sum_{u\in A,\,v\notin A}\pi(u)P(u,v).
\label{eq:cut_bound}
\end{equation}
Here $Q(A,A^c)$ is the stationary probability flow from one region to the other. Only transitions between the regions contribute to $\mathcal E(f)$, and reversibility makes the two directions equal \citep[Sec.~3.6]{aldous2002reversible}. When this flow is small relative to the regions' stationary masses, the bound places $\lambda_2$ near one.

For two rooms joined by a narrow doorway, this gives a mode that decays slowly. A large change across the rarely crossed doorway contributes little to its Dirichlet energy, while changes across frequent transitions within a room contribute more. When each room mixes quickly, this mode describes a contrast between the rooms, as in \cref{sec:theory}.

\noindent\textbf{What survives before mixing.}
Eventually, mixing removes these differences between starting states. To bound the remaining contrast, let $\rho=\max_{j\ge2}|\lambda_j|<1$. Completeness gives $\sum_{j\ge2}\psi_j(x)^2=\pi(x)^{-1}-1$, so Cauchy--Schwarz in \cref{eq:spectral_ratio} yields
\begin{equation}
\bigl|e^{G^*(x,y,\tau)}-1\bigr|
\le\rho^\tau\sqrt{\bigl(\pi(x)^{-1}-1\bigr)\bigl(\pi(y)^{-1}-1\bigr)}.
\label{eq:mixing_bound}
\end{equation}
Thus every ratio approaches one, $G^*\to0$, and progress between any two sources vanishes. When one positive mode decays more slowly than all the others, it determines the leading contrast. Specifically, if $\lambda_2>0$ and $\lambda_2>|\lambda_j|$ for every $j\ge3$, then applying $\log(1+z)=z+O(z^2)$ to \cref{eq:spectral_ratio} gives
\begin{equation}
G^*(x,y,\tau)=\lambda_2^\tau\psi_2(x)\psi_2(y)+o(\lambda_2^\tau).
\label{eq:second_mode}
\end{equation}
Without this separation, we must retain all dominant modes in the full expansion. Negative eigenvalues contribute terms that alternate in sign between odd and even horizons.

\subsection{What the embedding parameterization preserves}
\label{app:geometry_limits}
The learned score describes temporal relations through embedding alignment. When both embeddings have unit norm, we can express the same relation through their distance:
\begin{equation}
e^{G_\theta(x,y,\tau)}
=e^{\beta_0(\tau)+\beta_1(\tau)}
\exp\!\left[-\frac{\beta_0(\tau)}2\|h_\theta(x,\tau)-g_\theta(y,\tau)\|_2^2\right].
\label{eq:embedding_kernel}
\end{equation}
This identity holds for any learned coefficient $\beta_0(\tau)$ and offset $\beta_1(\tau)$. The coefficient controls how the score changes with embedding distance; its sign determines the direction of this dependence. At a fixed horizon, the offset rescales all exponentiated scores equally and leaves their ordering unchanged. Across horizons, these scales affect how much each horizon contributes to planning.

For tied encoders, comparison with the goal itself gives the exact identity
\begin{equation}
G_\theta(y_g,y_g,\tau)-G_\theta(x,y_g,\tau)
=\frac{\beta_0(\tau)}2\|h_\theta(x,\tau)-h_\theta(y_g,\tau)\|_2^2.
\label{eq:goal_anchor}
\end{equation}
In our evaluated tied-encoder models, the fitted alignment coefficients are positive and the offsets are negative. The positive coefficients make closer embeddings score higher and place the goal at a score maximum at each evaluated horizon, consistent with the geometric interpretation. This is an observation about the fitted models. A maximum at the goal does not rule out other local maxima or ensure that an improving move is available at every state.

Tied encoders also preserve a structural property of reversible walks. At a fixed horizon on a finite state space with $p_0(x)>0$, suppress the horizon argument and write
$K_{xy}=\exp(\beta_0\langle h_\theta(x),h_\theta(y)\rangle)$.
This kernel is symmetric for either sign of $\beta_0$. The normalized conditional matrix $p_0(y)K_{xy}/Z_\theta(x)$ is therefore reversible: with stationary weights proportional to $p_0(x)Z_\theta(x)$, the flow between $x$ and $y$ is proportional to $p_0(x)p_0(y)K_{xy}$, which is symmetric. This holds at each horizon, but does not require the learned conditionals to be powers of a single transition matrix.

These properties do not establish exact representation of every reversible walk. Tied unit embeddings give the same diagonal score at every state, whereas the walk's return ratio can vary from state to state.

\label{app:random_walk_proof}
There is a related feature construction for the ratio itself. For integers $k\ge1$, define the nonnegative features $f_k(x)_z=P^k(x,z)/\sqrt{\pi(z)}$. Detailed balance and matrix multiplication give $\langle f_k(x),f_k(y)\rangle=P^{2k}(x,y)/\pi(y)$. This diffusion-kernel construction \citep{coifman2006diffusion} represents the ratio as an inner product. Our affine inner product models its logarithm, so the construction does not establish exact representation by our model.

\subsection{Planning: GP and PAP as temperature limits}
\label{app:planning}
\label{app:variational}
\label{sec:planning_theory}
The same action can improve the goal score at one horizon and reduce it at another. The temperature objective lets each action place more weight on horizons where it makes progress, while penalizing departures from uniform weights.

Fix a goal $y_g$, a nonempty finite horizon set $\mathcal T$, and finite scores. For a candidate action $a$, write
$\Delta_\tau(a)=e^{G_\theta(\widehat x_{t+1}^{\,a},y_g,\tau)}-e^{G_\theta(x_t,y_g,\tau)}$.
Let $U$ be the uniform distribution on $\mathcal T$, and let $\mathcal P(\mathcal T)$ be the set of distributions over these horizons.

\begin{proposition}[GP and PAP as temperature limits]
\label{prop:horizon_bridge}
For $\beta>0$, the planning objective satisfies
\begin{equation}
\begin{aligned}
V_\beta(a)&:=\beta\log\!\left[\frac1{|\mathcal T|}\sum_\tau e^{\Delta_\tau(a)/\beta}\right]\\
&=\max_{q\in\mathcal P(\mathcal T)}\left\{\sum_\tau q(\tau)\Delta_\tau(a)-\beta\,\mathrm{KL}(q\|U)\right\},
\end{aligned}
\label{eq:horizon_bridge}
\end{equation}
with unique maximizer $q_\beta(\tau|a)\propto e^{\Delta_\tau(a)/\beta}$. As $\beta\to0^+$, $V_\beta(a)$ converges to $\max_\tau\Delta_\tau(a)$ (GP); as $\beta\to\infty$ it converges to $|\mathcal T|^{-1}\sum_\tau\Delta_\tau(a)$ (PAP).
\end{proposition}
\begin{proof}
With $\mathrm{KL}(q\|U)=\sum_\tau q(\tau)\log(|\mathcal T|q(\tau))$ and $0\log0=0$, direct substitution gives
\begin{equation}
\sum_\tau q(\tau)\Delta_\tau(a)-\beta\,\mathrm{KL}(q\|U)
=V_\beta(a)-\beta\,\mathrm{KL}(q\|q_\beta(\cdot|a)).
\label{eq:gibbs_identity}
\end{equation}
The right-hand side is largest only when $q=q_\beta$, because KL is nonnegative and vanishes only when the two distributions agree. This proves the variational identity and uniqueness. To obtain the limits, write $M=\max_\tau\Delta_\tau(a)$. Evaluating the variational objective at weights concentrated on a maximizing horizon, at $U$, and at $q_\beta$ gives
\begin{equation}
\begin{aligned}
M-\beta\log|\mathcal T|&\le V_\beta(a)\le M,\\
\frac1{|\mathcal T|}\sum_\tau\Delta_\tau(a)&\le V_\beta(a)\le\sum_\tau q_\beta(\tau|a)\Delta_\tau(a).
\end{aligned}
\label{eq:sandwich}
\end{equation}
The first line gives the GP limit; the second gives the PAP limit because $q_\beta\to U$ as $\beta\to\infty$.
\end{proof}
At low temperature, an action is judged by the horizon where it makes the most progress. At high temperature, all horizons receive equal weight, so gains and losses offset each other. Multiplying all progress values by the same positive constant preserves GP and PAP rankings. At finite temperature, preserving the ranking also requires scaling $\beta$ by that constant.

\noindent\textbf{Potential ascent and revisits.}
\label{app:progress}
PAP evaluates every move through one function of the state. For the fixed goal, define
\begin{equation}
\Phi(x)=\frac1{|\mathcal T|}\sum_{\tau\in\mathcal T}e^{G_\theta(x,y_g,\tau)}.
\label{eq:potential}
\end{equation}
PAP progress is exactly $\Phi(\widehat x_{t+1}^{\,a})-\Phi(x_t)$. Since the current-state term is shared by all actions, PAP ranks candidates by their next-state potential. If the model, goal, and horizon set stay fixed, predictions match deterministic outcomes, and every executed step has strictly positive progress, then $\Phi(x_{t+1})>\Phi(x_t)$. Returning to a previous state would restore its previous potential, which is impossible. The argument does not ensure that an improving action exists at every non-goal state.

GP can favor different horizons on successive moves and need not
increase a common state potential. Consider two non-goal states
with exponentiated score vectors $(0.2,0.8)$ and $(0.8,0.2)$ across
two horizons. Each state offers a move to the other and a stay
action. Moving improves one component by $0.6$ and reduces the
other by $0.6$, so GP prefers moving in both directions and cycles
between the states. At every finite temperature, moving also gives
$V_\beta=\beta\log\cosh(0.6/\beta)>0$, while staying gives zero.
PAP assigns both states potential $0.5$, so neither move has
strictly positive PAP progress. This cycling example is realizable
even with tied, positive unit embeddings,\footnote{Choose positive
unit vectors $u,v$ with $\langle u,v\rangle=1/2$, use $u$ for the
goal at both horizons, and assign $(v,u)$ and $(u,v)$ to the two
states across horizons. Setting $\beta_0=\log16$ and
$\beta_1=\log0.05$ gives the stated exponentiated scores.}
showing that these embedding constraints do not prevent GP from
cycling. The scores need not be calibrated temporal ratios of a
common walk, so the example concerns the learned-score planner.

\subsection{Geometric weights and goal reaching}
\label{app:geometric}
With exact transition ratios, the potential has a direct interpretation under the exploration walk: $|\mathcal T|p_0(y_g)\Phi(x)$ is the expected number of visits to $y_g$ at horizons in $\mathcal T$. Counting visits differs from measuring whether the goal is reached, since one trajectory may visit it several times.

Geometric weights let us relate these two quantities. Consider any finite Markov chain $P$, take a fixed $p_0(y_g)>0$, and let $0<\gamma<1$. Weighting all horizons, including zero, defines
\begin{equation}
\Phi_\gamma(x)=(1-\gamma)\sum_{\tau=0}^\infty\gamma^\tau\frac{P^\tau(x,y_g)}{p_0(y_g)}.
\label{eq:geometric_potential}
\end{equation}
This potential measures discounted occupancy of the goal relative to $p_0(y_g)$. The following result relates it to the time of first arrival. Neither reversibility nor an absorbing goal is required.

\begin{proposition}[Geometric weights connect occupancy to first arrival]
\label{prop:geometric_goal}
Let $T_g=\inf\{t\ge0:X_t=y_g\}$ be the first arrival time under $P$, with $\gamma^\infty=0$, and let $\E_x$ denote expectation for the walk started at $x$. Then
\begin{equation}
\Phi_\gamma(x)=\E_x[\gamma^{T_g}]\,\Phi_\gamma(y_g).
\label{eq:geometric_hitting}
\end{equation}
Every non-goal state from which the goal is reachable satisfies
\begin{equation}
\max_{y:P(x,y)>0}\Phi_\gamma(y)\ge\Phi_\gamma(x)/\gamma>\Phi_\gamma(x).
\label{eq:geometric_progress}
\end{equation}
\end{proposition}
\begin{proof}
There are no visits to the goal before $T_g$. Once the walk arrives, its subsequent visits have the same law as those of a walk started at the goal. The strong Markov property therefore gives
\[
\E_x\!\left[\sum_{t\ge0}\gamma^t\mathbf1\{X_t=y_g\}\right]
=\E_x[\gamma^{T_g}]\,
\E_{y_g}\!\left[\sum_{j\ge0}\gamma^j\mathbf1\{X_j=y_g\}\right].
\]
Multiplying by $(1-\gamma)/p_0(y_g)$ proves \cref{eq:geometric_hitting} \citep[Lemma~2.25]{aldous2002reversible}. To show that an improving move exists, split off the horizon-zero term:
$\Phi_\gamma(x)=\frac{1-\gamma}{p_0(y_g)}\mathbf1\{x=y_g\}+\gamma\sum_yP(x,y)\Phi_\gamma(y)$.
At a reachable non-goal state, $\Phi_\gamma(x)>0$ and the indicator vanishes. The average next-state potential is therefore $\Phi_\gamma(x)/\gamma>\Phi_\gamma(x)$. At least one successor with $P(x,y)>0$ has value at least this average, proving \cref{eq:geometric_progress}.
\end{proof}
This gives a goal-reaching guarantee under exact local control. Suppose all walk-supported moves are available as deterministic actions, their outcomes are predicted exactly, and planning stops at the goal. From any start that can reach the goal under $P$, choosing a successor with largest $\Phi_\gamma$ strictly increases the potential. Its value remains positive, so the goal remains reachable. No state can be revisited, and the trajectory cannot stop at a non-goal state. It must therefore reach the goal in at most $|\mathcal S|-1$ moves. The known horizon-zero term anchors the goal value; omitting it can destroy this guarantee.

The geometric aggregate is proportional to discounted occupancy, the quantity connected to contrastive goal-conditioned RL \citep{eysenbach2022crl}. Dividing by its value at the goal gives the discounted first-arrival value. Under the exact deterministic control assumptions above, choosing the successor with largest value is a policy-improvement step over the exploration behavior. Our planner uses a finite set of uniformly weighted positive horizons, which does not generally satisfy this Bellman identity. The geometric extension therefore explains a connection to goal reaching.

\newpage
\section{Algorithms}
\label{app:algorithms}
\Cref{alg:learning,alg:dynamics_learning} learn the temporal representation and local dynamics separately. \Cref{alg:planning} uses both models for closed-loop planning, with their parameters fixed.

\newcounter{paperalgorithm}
\newcounter{paperalgline}
\begingroup
\newenvironment{paperalgsteps}{%
  \begin{list}{\arabic{paperalgline}.}{%
    \usecounter{paperalgline}%
    \setlength{\leftmargin}{1.5em}%
    \setlength{\labelwidth}{1.2em}%
    \setlength{\labelsep}{.3em}%
    \setlength{\itemsep}{4pt}%
    \setlength{\parsep}{0pt}%
    \setlength{\topsep}{4pt}}%
}{\end{list}}
\noindent
\begin{minipage}{\linewidth}
\normalsize\raggedright
\hrule\vspace{4pt}
\refstepcounter{paperalgorithm}\label{alg:learning}%
\textbf{Algorithm \thepaperalgorithm: Learning $\theta$}\par
\vspace{4pt}\hrule\vspace{5pt}
\textbf{Input:} Exploration trajectories, $\mathcal T$, $p(\tau)$, reference sampler $p_0$, and negative count $N$.\par
\begin{paperalgsteps}
\item Initialize temporal parameters $\theta$.
\item \textbf{while} temporal updates remain \textbf{do}
\item \hspace*{1em}Sample a horizon $\tau\sim p(\tau)$.
\item \hspace*{1em}Sample a batch of pairs $(x,y)$ from $p_{\rm data}(x,y|\tau)$ within episodes.
\item \hspace*{1em}For each source $x$, sample $N$ independent negative targets from $p_0$.
\item \hspace*{1em}Compute $G_\theta(x,y,\tau)-\log N$ for positive and negative pairs.
\item \hspace*{1em}Update $\theta$ by \cref{eq:nce}: average the positive loss plus the sum of negative losses over sources.
\item \textbf{end while}
\item \textbf{return} learned parameters $\theta$.
\end{paperalgsteps}
\hrule
\end{minipage}
\par\vspace{18pt}
\noindent\begin{minipage}{\linewidth}
\normalsize\raggedright
\hrule\vspace{4pt}
\refstepcounter{paperalgorithm}\label{alg:dynamics_learning}%
\textbf{Algorithm \thepaperalgorithm: Learning $\phi$}\par
\vspace{4pt}\hrule\vspace{5pt}
\textbf{Input:} Action-labeled one-step transitions with distribution $p_{\rm step}$.\par
\begin{paperalgsteps}
\item Initialize local-dynamics parameters $\phi$.
\item \textbf{while} dynamics updates remain \textbf{do}
\item \hspace*{1em}Sample a batch of transitions $(x_t,a_t,x_{t+1})\sim p_{\rm step}$.
\item \hspace*{1em}Update $\phi$ by \cref{eq:dynloss} for state prediction.
\item \textbf{end while}
\item \textbf{return} learned parameters $\phi$.
\end{paperalgsteps}
\hrule
\end{minipage}
\par\vspace{18pt}
\noindent\begin{minipage}{\linewidth}
\normalsize\raggedright
\hrule\vspace{4pt}
\refstepcounter{paperalgorithm}\label{alg:planning}%
\textbf{Algorithm \thepaperalgorithm: Planning}\par
\vspace{4pt}\hrule\vspace{5pt}
\textbf{Input:} Frozen $G_\theta$ and $F_\phi$, goal $y_g$, horizon set $\mathcal T$, temperature $\beta>0$ (or either limit), and action budget.\par
\begin{paperalgsteps}
\item Observe the initial $x_0$ and set $t=0$.
\item \textbf{while} goal not reached and budget remains \textbf{do}
\item \hspace*{1em}Construct candidate set $\mathcal A(x_t)$.
\item \hspace*{1em}\textbf{for each} $a\in\mathcal A(x_t)$ \textbf{do}
\item \hspace*{2em}Predict $\widehat x_{t+1}^{\,a}=F_\phi(x_t,a)$.
\item \hspace*{2em}Compute progress for each $\tau\in\mathcal T$:
  $\Delta_\tau(a)=e^{G_\theta(\widehat x_{t+1}^{\,a},y_g,\tau)}-e^{G_\theta(x_t,y_g,\tau)}$.
\item \hspace*{1em}\textbf{end for}
\item \hspace*{1em}Select $a_t^*$ by \cref{eq:unified_plan}, using maximum progress as $\beta\to0^+$ or mean progress as $\beta\to\infty$.
\item \hspace*{1em}Execute only $a_t^*$ in the environment.
\item \hspace*{1em}Receive the actual $x_{t+1}$ and set $t\leftarrow t+1$.
\item \textbf{end while}
\end{paperalgsteps}
\vspace{2pt}\hrule
\end{minipage}
\par
\endgroup
\par

\clearpage
\noindent\textbf{PyTorch-Style Pseudocode}\par\smallskip
The compact version below shows one update for each model and one planning decision. \texttt{pair\_score} returns a $B\times B$ matrix of temporal scores at a shared lag; \texttt{goal\_score} returns candidate-by-horizon scores. Here \texttt{F} denotes \texttt{torch.nn.functional}. Planning uses every integer horizon from 1 to $\tau_{\max}$.

\begin{lstlisting}[style=paperpython]
def train_temporal(current, future, tau):
    logits = temporal.pair_score(current, future, tau)
    logits = logits - math.log(len(current) - 1)
    positives = logits.diagonal()
    negatives = F.softplus(logits).sum(1)
    negatives = negatives - F.softplus(positives)
    loss = (F.softplus(-positives) + negatives).mean()
    temporal_opt.zero_grad()
    loss.backward()
    temporal_opt.step()

def train_dynamics(state, action, next_state):
    prediction = dynamics(state, action)
    error = (prediction - next_state).square()
    loss = error.flatten(1).sum(1).mean()
    dynamics_opt.zero_grad()
    loss.backward()
    dynamics_opt.step()

@torch.no_grad()
def choose_action(state, goal, actions, taus, planner):
    states = state.expand(len(actions), -1)
    next_states = dynamics(states, actions)
    here = temporal.goal_score(state, goal, taus).exp()
    there = temporal.goal_score(next_states, goal, taus).exp()
    progress = there - here
    if planner == "GP":
        value = progress.amax(1)
    else:
        value = progress.mean(1)
    return actions[value.argmax()]
\end{lstlisting}
At test time, both models are in evaluation mode. Execute the returned action, observe the new state, and repeat. The dynamics call illustrates learned state prediction; oracle experiments instead obtain candidate outcomes from the simulator.

\clearpage
\FloatBarrier
\section{Experiments}
\label{app:experiments}
\label{app:protocol}
\label{app:provenance}
\label{app:setup}

\FloatBarrier

\subsection{Representation Probes}

\begin{figure}[!htbp]
\centering
\centering\includegraphics[width=\linewidth]{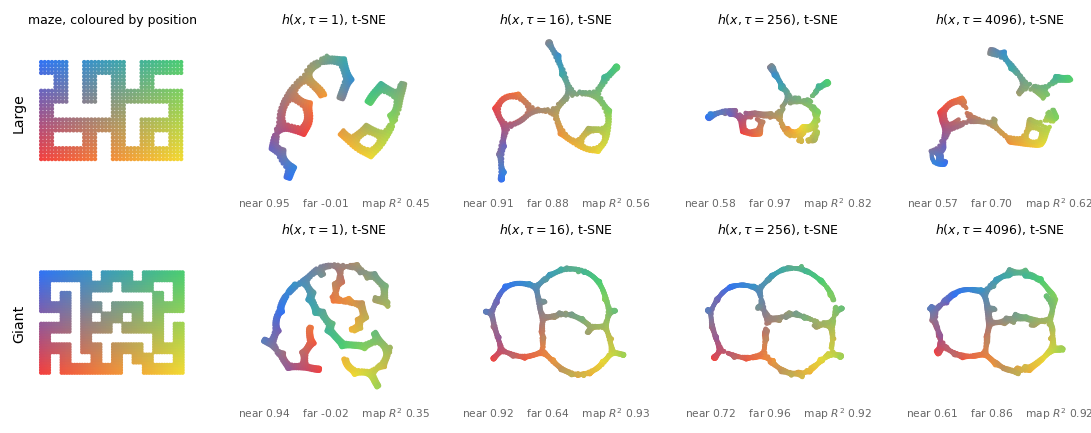}
\caption{Learned embeddings $h(x,\tau)$ on the Large (top) and Giant (bottom) state mazes; the left column colours each cell by its position and the other columns keep those colours. Every panel is rotated and reflected onto the maze by Procrustes, since t-SNE leaves the orientation undetermined.}
\label{fig:latent}
\end{figure}

\paragraph{Representation Probes and Visualizations.}
\label{app:representation}
\Cref{fig:latent,tab:latent_rank} compare embedding distances with maze geometry at different horizons. Near pairs are at most eight geodesic steps apart. Each t-SNE panel is fitted separately and aligned to the maze by an orthogonal Procrustes transform; projection coordinates are not physical distances.

\begin{table}[!htbp]
\centering\small
\caption{Spearman correlation between embedding distance and geodesic distance for near ($\le 8$ steps) and far pairs, Large maze.}
\label{tab:latent_rank}
\begin{tabular}{@{}lcccc@{}}
\toprule
$\tau$ & 1 & 16 & 256 & 4096\\
\midrule
near & \textbf{0.95} & 0.91 & 0.58 & 0.57\\
far  & $-$0.01 & 0.88 & \textbf{0.97} & 0.70\\
\bottomrule
\end{tabular}
\end{table}

\begin{figure}[!htbp]
\centering
\includegraphics[width=\linewidth]{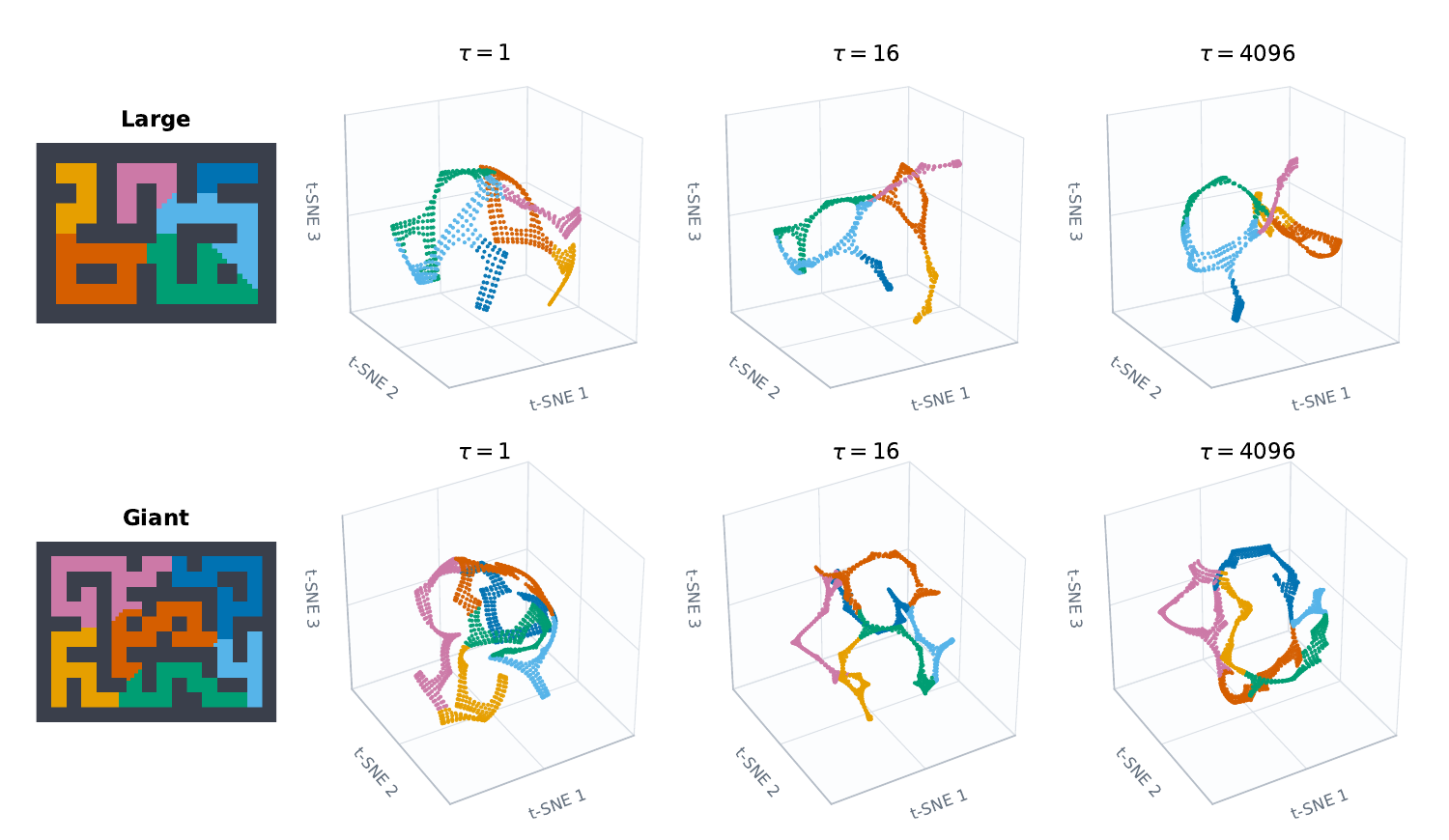}
\caption{3D t-SNE of state embeddings on Large (top) and Giant (bottom). At $\tau=1$, 3D views reveal connectivity obscured in 2D. Colours match the maze regions at left; axes are not physical coordinates.}
\label{fig:tsne_3d}
\end{figure}

\paragraph{Three-dimensional views.}
\Cref{fig:tsne_3d} supplements the Giant-maze 2D projections in \cref{fig:overview}B with 3D t-SNE views of both mazes at $\tau=1$, $16$, and $4096$. We use perplexity 30, PCA initialization, and random seed 0, fitting each maze and horizon separately. Rotation/reflection and uniform scaling align each fit to the maze plane without flattening its third dimension. Within each maze, all three panels share a camera angle and coordinate scale. The projections illustrate neighbourhood structure, not physical distances or comparable coordinates across horizons.

\FloatBarrier
\subsection{Demonstrations}

\paragraph{Planning demonstrations.}
\Cref{fig:overview}C uses all integer horizons from 1 to 512 on Large (tasks 1 and 3) and from 1 to 2048 on Giant (tasks 1 and 2), with one-step lookahead and execution (0.2 units). These illustrative rollouts use different horizon ranges and a different planning radius from the aggregate evaluations in \cref{tab:maze}. For GP, path colour marks the horizon with the largest improvement for the chosen action; for PAP, it marks the horizon contributing most to the chosen candidate's pooled score, not a separately selected planning horizon. The recorded values are not temporally smoothed. Matching marker colours pair each start with its recorded episode goal.

\begin{figure}[!htbp]
\centering
\includegraphics[width=\textwidth]{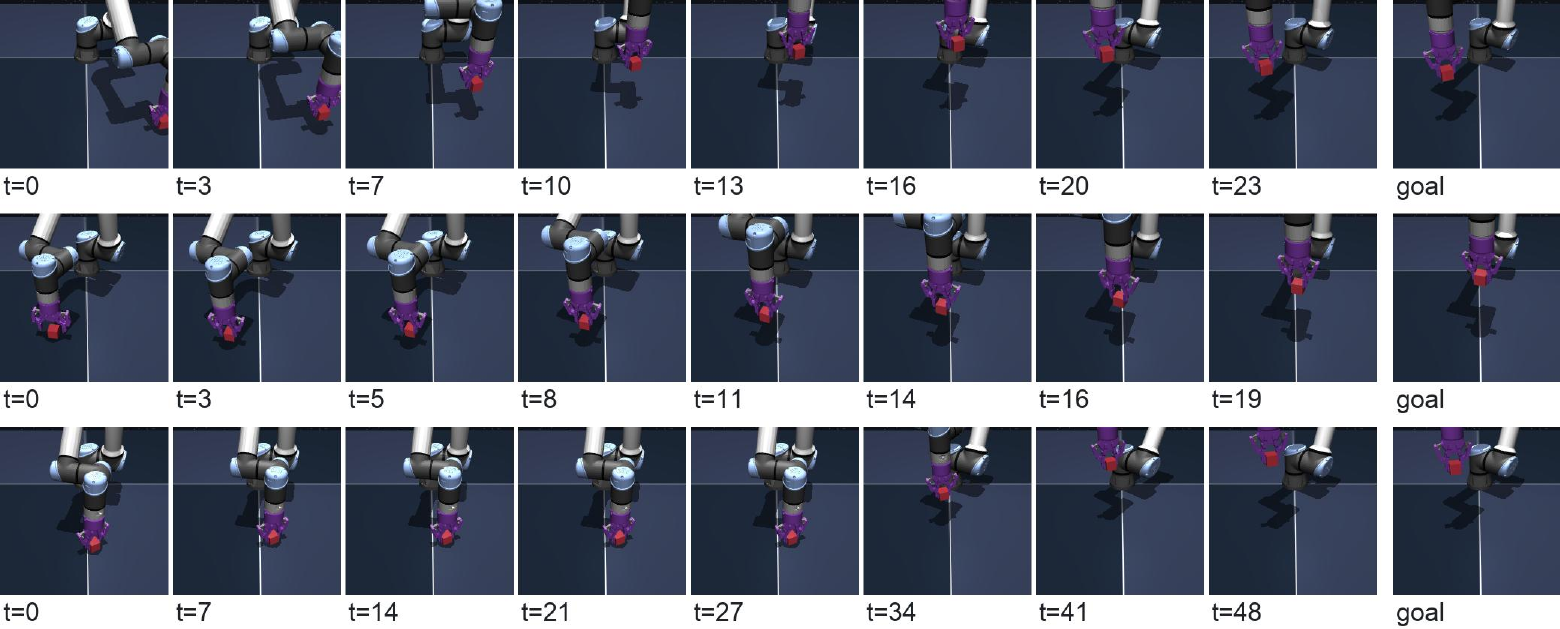}
\caption{Three held-out cube-single episodes planned from pixels, one per row, each showing eight evenly spaced executed steps with the goal image last. The cube travels 0.455, 0.277, and 0.325\,m, respectively; in every row the planner reaches the cube, grasps it, and carries it to the goal.}
\label{fig:cube_rollout}
\end{figure}

\paragraph{Manipulation demonstrations.}
\Cref{fig:cube_rollout} shows three planned episodes of different lengths, with evenly spaced frames followed by the goal image.

\FloatBarrier

\end{document}